\documentclass[runningheads]{llncs}
\usepackage{graphicx}%
\usepackage{hyperref}
\usepackage{enumitem}
\usepackage{verbatim}
\usepackage{xcolor}
\usepackage{amssymb}
\usepackage{amsmath}
\usepackage{array}
\usepackage{pifont}
\usepackage{tikz}
\usepackage{booktabs}
\usepackage[T1]{fontenc}%
\newcolumntype{R}{>{\centering\arraybackslash}m{1cm}}
\newif\ifshowcomments
\showcommentsfalse
\ifshowcomments
  \newcommand{\JC}[1]{{\color{green}{\bf JC: #1}}}

\else
  \newcommand{\JC}[1]{}

\fi
\usepackage{fancyvrb}
\definecolor{verbgray}{rgb}{0.3, 0.3, 0.3}
\DefineVerbatimEnvironment{myverbatim}{Verbatim}{fontfamily=cmtt,formatcom=\color{verbgray}}

\definecolor{dred}{rgb}{127,0,0}
\definecolor{dblue}{rgb}{0,0,178}

\newcommand{\showstd}[1]{{\textcolor{gray}{#1}}}

\newcommand{\myparagraph}[1]{\noindent\textbf{#1.}\enspace}

\newcommand{\best}[1]{\textbf{#1}}
\newcommand{\bestci}[1]{\textbf{#1}}

\begin{document}
\title{Learning to Read Where to Look: Disease-Aware Vision--Language Pretraining for 3D CT}
\titlerunning{Disease-Aware Vision--Language Pretraining for 3D CT}
\author{
Simon Ging\inst{1,2} \and 
Philipp Arnold\inst{3} \and
Sebastian Walter\inst{4} \and
Hani Alnahas\inst{1} \and
\\
Hannah Bast\inst{4} \and
Elmar Kotter\inst{3} \and
Jiancheng Yang\inst{5,6} \and
\\
Behzad Bozorgtabar\inst{2} \and
Thomas Brox\inst{1}\\
\email{gings@cs.uni-freiburg.de}\\
\url{https://radfinder.github.io/}
}
\authorrunning{S. Ging et al.}
\institute{
Computer Vision Group, University of Freiburg, Germany
\and
Adaptive \& Agentic AI (A3) Lab, Aarhus University, Denmark
\and 
Department of Radiology, Medical Center -- University of Freiburg, Germany
\and
Chair of Algorithms and Data Structures, University of Freiburg, Germany
\and
ELLIS Institute Finland
\and
School of Electrical Engineering, Aalto University, Finland
}
\maketitle
\begin{abstract}%
Recent 3D CT vision--language models align volumes with reports via contrastive pretraining, but typically rely on limited public data and provide only coarse global supervision.
We train a 3D CT vision--language model on 98k report--volume pairs (50k patients) collected at a single hospital, combined with public datasets, using SigLIP-style contrastive pretraining together with prompt-based disease supervision in the shared vision--text embedding space.
On CT-RATE, our model achieves state-of-the-art text-to-image retrieval (R@10 31.5 vs.\ 22.2) and competitive disease classification (AUC 83.8 vs.\ 83.8), with consistent results on Rad-ChestCT (AUC 77.0 vs.\ 77.3).
We further observe that radiologists routinely reference specific images within their reports (e.g., ``series~X, image~Y''), linking textual descriptions to precise axial locations.
We automatically mine 262k such snippet--slice pairs and introduce the task of intra-scan snippet localization---predicting the axial depth referred to by a text snippet---reducing mean absolute error to 36.3\,mm at 12\,mm feature resolution, compared with 67.0\,mm for the best baseline.
Adding this localization objective leaves retrieval and classification broadly unchanged within confidence bounds, yielding a single unified model for retrieval, classification, and intra-scan grounding.
\end{abstract}
\section{Introduction}

The volume of radiological imaging continues to grow, with CT examinations increasing steadily, while the number of radiologists has not kept pace~\cite{rimmer2017radiologist}.
Each CT study requires careful analysis: a radiologist reviews hundreds of axial slices, identifies findings, and dictates a free-text report.
Foundation models that learn general-purpose representations from large-scale imaging data offer a path toward assisting radiologists across diverse tasks---from automated retrieval and triage to report generation---without task-specific annotation.

Recent 3D CT vision--language models (VLMs) learn such representations by aligning CT volumes with radiology reports via contrastive pretraining.
CT-CLIP~\cite{ctrate} pioneered this direction on the public CT-RATE dataset of 50k chest CT volumes.
SPECTRE~\cite{spectre} scaled self-supervised and cross-modal pretraining to 230k volumes and achieved strong retrieval results, while other approaches leverage structured entity extraction~\cite{mpsct} or hierarchical vision architectures~\cite{pillar0}.
These methods align entire volumes with entire reports, providing only coarse, global supervision.
Other approaches improve supervision granularity by aligning at the organ level: MedVista3D~\cite{medvista3d} and fVLM~\cite{fvlm} both use an external segmentation model to identify anatomical regions and contrast each region's visual features against LLM-generated region descriptions.
While effective, this requires a pretrained segmentation model and curated region-level text that may not capture the specific findings a radiologist highlights.

In this work, we train \textit{RadFinder}, a 3D CT VLM on 98k report--volume pairs (50k patients) from a single hospital, combined with public datasets.
The model uses SigLIP~\cite{siglip} contrastive pretraining on full reports with pretrained medical vision and text encoders, and integrates structured disease labels as text prompts within the contrastive objective (Sec.~\ref{sec:prompt_training}).
RadFinder achieves state-of-the-art text-to-image retrieval and competitive disease classification on CT-RATE and Rad-ChestCT~\cite{radchestct}, demonstrating strong cross-dataset transfer.

We additionally observe that radiology reports contain a largely untapped form of local supervision: radiologists frequently reference specific slices within a scan---for example, ``\textit{hepatic lesion, see series~4, image~38}''.
We mine 262k such snippet--slice pairs and propose the task of \emph{intra-scan snippet localization}: given a text snippet, predict the axial depth it refers to within a CT volume.
While slice-level matching from reports has been explored at small scale~\cite{huemann2025}, we formulate this in 3D feature space and achieve results substantially better than simple baselines.
Training localization jointly with global objectives does not degrade retrieval or classification, yielding a unified model for all three tasks.
\\\\
\myparagraph{Contributions}
\begin{itemize}[leftmargin=*,itemsep=2pt,topsep=2pt]
    \item We build a large-scale 3D report--volume dataset (98k pairs, 50k patients) from clinical routine at a single hospital and automatically mine 262k snippet--slice pairs that provide weak local supervision without additional annotation.
    \item Using this local supervision signal, we propose the task of intra-scan snippet localization, establish baselines, and show that localization can be trained jointly with global objectives without degrading retrieval or classification.
    \item We train a 3D CT VLM by combining contrastive pretraining on reports with prompt-based disease label training and snippet localization. Our model achieves state-of-the-art retrieval and competitive disease classification on external benchmarks, demonstrating strong cross-dataset transfer. Code and pretrained models will be made publicly available.
\end{itemize}

\section{Dataset}

We collect 97,760 report--volume pairs from 50,474 patients at a single hospital, spanning 13~years of clinical practice.
The dataset covers chest (46\%), abdomen (22\%), and combined chest--abdomen (33\%) studies.
The radiologist report for the study contains detailed findings and a summary impression, and we select the single largest axial series from each study.
In-plane resolution and slice thickness medians are 0.71\,mm and 3.0\,mm, respectively.

\myparagraph{Snippet mining}
Radiologists frequently reference specific slices when dictating reports (e.g., ``hepatic lesion, see series~4, image~38'' or ``pulmonary nodule in the right lower lobe (3/72)'').
We extract these series/image references via pattern matching heuristics.
Each snippet associates a short textual finding with a precise axial position in the scan, providing weak local supervision without additional annotation.
To verify our pipeline, we compare against manually extracted references from 100 reports and find that our heuristics achieve 99.4\% precision and 90.2\% recall (F1 94.6\%).
To ensure correct spatial alignment, we keep only slices where slicing the 3D volume at the referenced position produces identical content to the original 2D image file stored by the scanner.
After filtering, this yields 261,800 snippet--slice pairs across all scans, 2.7 per scan on average.

\myparagraph{Text processing}
We anonymize all text by removing patient and physician identifiers via rule-based matching and converting absolute dates into relative references (e.g., ``Spine injury 5 years ago'').
Reports are translated from German to English using Gemma~3~27B~\cite{gemma3}; all training is conducted on the English translations.
We apply the RATE protocol from Pillar-0~\cite{pillar0}, using LLM-based question answering to classify each report into 93 chest and 226 abdomen binary findings across 30 organ categories; these structured labels serve as input for prompt-based training (Sec.~\ref{sec:prompt_training}).
Data is split 80/10/10 by randomized patient IDs, with no patient overlap with public evaluation datasets.

\section{Method}

\begin{figure}[ht]
\makebox[\textwidth][c]{%
\includegraphics[width=1.0\textwidth]{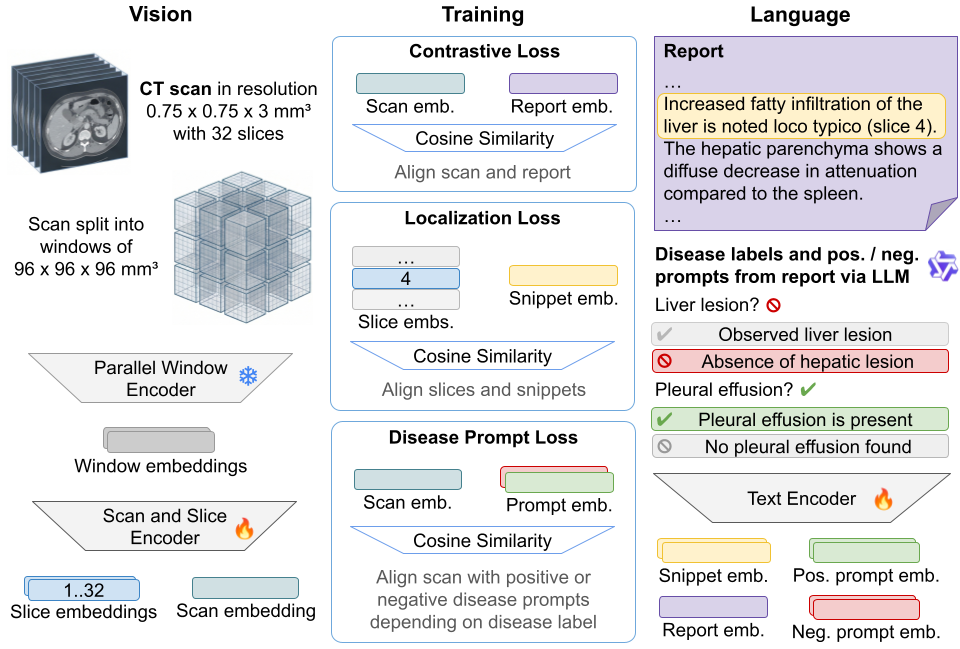}
}
\caption{
Overview of the RadFinder architecture and training pipeline. \textbf{Left (Vision):} A 3D CT scan is processed by a frozen parallel window encoder, followed by a trainable encoder to extract both a global volume embedding and local slice-level embeddings. \textbf{Right (Language):} A trainable text encoder processes full radiology reports, localized text snippets (e.g., the sentence that references the fourth slice out of the 32 slices in this example scan), and LLM-extracted positive/negative disease prompts. \textbf{Center (Training):} The model is optimized via three contrastive objectives in a shared embedding space: a global Contrastive Loss aligning the full scan and report, an intra-scan Localization Loss aligning text snippets with their specific slices, and a Disease Prompt Loss aligning the global scan with corresponding disease prompt descriptions
}
\label{fig1}
\end{figure}

\subsection{Global Vision--Language Pretraining}

We initialize from the pretrained SPECTRE~\cite{spectre} model, which uses a two-stage vision encoder: a ViT-Large local backbone processes each volume in separate $128 \times 128 \times 32$-voxel windows at $0.75 \times 0.75 \times 3.0$\,mm$^3$ spacing, and a 4-layer global feature combiner aggregates the window-level representations.
The text encoder is Qwen3-Embedding~\cite{qwen3} (0.6B) with LoRA adapters.
SigLIP~\cite{siglip} projection heads map both modalities to a 512-dim shared space.
We freeze the local vision backbone and fine-tune all other modules with SigLIP contrastive loss.
Unlike SPECTRE, which crops to a fixed grid, we process full volumes with variable input shapes.
We train on up to four datasets totaling 159k report--volume pairs: \textit{RefCT} (internal, 78k in the training split), CT-RATE~\cite{ctrate} (47k), Merlin~\cite{merlin} (15k), and INSPECT~\cite{inspect} (19k).
During training, we apply three text augmentations, each with probability~0.2: replacing the full report with concatenated organ-level descriptions from the RATE pipeline~\cite{pillar0}, removing historic comparisons (``compared to prior exam\ldots'') via LLM parsing following CT-RATE~\cite{ctrate}, or dropping the \textit{findings} section.

\subsection{Prompt-Based Disease Label Training}
\label{sec:prompt_training}

To improve text-based disease classification, we augment the contrastive loss with disease label prompts.
During training, labels are represented as text prompts and supervised via a BCE loss on the cosine similarity difference between the volume embedding and positive/negative prompt embeddings.
We do not train a separate classification head; supervision is applied entirely in the shared vision–text embedding space.

\myparagraph{Labels} Using the RATE protocol~\cite{pillar0}, we extract binary findings from each report via LLM-based question answering with Qwen3-30B-A3B~\cite{qwen3} (93 chest and 226 abdomen findings).
We additionally derive binary labels for the 18 CT-RATE disease classes by mapping the 93 RATE chest findings.

\myparagraph{Prompts} For each finding~$q$, we construct three positive and three negative text prompt variants (e.g., ``Pleural effusion is present.'', ``No pleural effusion is identified.'').
During training, one variant is sampled randomly; at inference, the three embeddings are averaged.

\myparagraph{Loss}
Let $\mathbf{z}$ denote the L2-normalized image embedding and $\mathbf{p}_q^{+}, \mathbf{p}_q^{-}$ the positive and negative prompt embeddings for finding~$q$.
We classify each finding via the logit difference scaled by the shared SigLIP temperature~$\tau$:

\begin{align}
\mathcal{L}_{\text{prompt}}
= \frac{1}{|\mathcal{M}|}\sum_{q\in\mathcal{M}}
w_q\Big(
-\alpha_q\, y_{q}\,\log\sigma(x_{q})
-(1- y_{q})\log\!\bigl(1-\sigma(x_{q})\bigr)
\Big)
\label{eq:promptloss}
\\
x_{q} = (\mathbf z^\top \mathbf p_{q}^{+}-\mathbf z^\top \mathbf p_{q}^{-}) / \tau,
\qquad 
\alpha_q=\min(n^+_q / n^-_q, 20)
\label{eq:promptdefs}
\end{align}

where $y_{q} \in \{0,1\}$ is the binary finding label, $\sigma$ is the logistic sigmoid, $\mathcal{M}$ is the set of valid image--question pairs, $w_q$ upweights the 18 CT-RATE classes to balance them against the larger RATE label set, and $\alpha_q$ handles per-question label imbalance given $n^+_q$ positive and $n^-_q$ negative training examples.
The prompt loss is weighted by $\lambda$ relative to the SigLIP loss.

The idea of using label text as prompts in contrastive training originates in CXR-CLIP~\cite{cxrclip} and UniCL~\cite{unicl}.
The closest prior work in 3D CT either extracts free-text disease descriptions from reports with GPT-4 for contrastive training (BrgSA~\cite{brgsa}) or trains an explicit classification head on LLM-extracted labels before contrastive alignment (MPS-CT~\cite{mpsct}).
Our approach uses structured binary labels directly as text prompts within the contrastive objective.

\subsection{Intra-Scan Localization Loss}

\label{sec:localization_loss}

To encourage depth-aware representations, we introduce an intra-scan localization objective.
Given a text snippet describing a radiological finding and the corresponding CT volume,
the model classifies which depth position along the axial axis the snippet refers to.
Crucially, negatives come from other depth positions \emph{within the same scan},
avoiding cross-sample false negatives that arise when different patients share similar pathology.

\myparagraph{Depth features}
Let $\mathbf{z}_d \in \mathbb{R}^E$ denote the L2-normalized image embedding at depth position $d \in \{1, \dots, D\}$, with one position covering 12mm in the axial plane,
obtained by processing backbone patch features through the global feature combiner,
averaging over the coronal and sagittal dimensions, and projecting through the SigLIP head.
Let $\mathbf{t} \in \mathbb{R}^E$ denote the L2-normalized projected text embedding of the snippet.

\myparagraph{Gaussian soft target}
Each snippet is associated with a specific axial slice, whose physical position determines
a ground-truth depth index $d^* \in \{1, \dots, D\}$ in the feature grid.
Since annotations are inherently imprecise---a finding typically extends to neighboring slices---we
define a one-hot indicator $\mathbf{m} \in \{0,1\}^D$ with $m_{d^*} = 1$
and convolve it with a normalized 1-D Gaussian kernel $g$ of standard deviation $\sigma{=}2$:
\begin{equation}
    \tilde{\mathbf{m}} = \frac{(\mathbf{m} * g)}{\| \mathbf{m} * g \|_1},
    \qquad
    g_k = \frac{1}{Z} \exp\!\Bigl(-\frac{k^2}{2\sigma^2}\Bigr),
    \quad |k| \leq \lceil 3\sigma \rceil.
    \label{eq:soft_target}
\end{equation}

\myparagraph{Loss}
We compute cosine-similarity logits between the snippet embedding and each depth feature,
scaled by a temperature $\tau{=}0.1$, and minimize the cross-entropy against the soft target $\tilde{\mathbf{m}}$:
\begin{equation}
    \ell_d = \mathbf{z}_d^\top \mathbf{t} /\tau,
    \qquad
    \mathcal{L}_{\text{loc}} = -\sum_{d=1}^{D} \tilde{m}_d \, \log \frac{\exp(\ell_d)}{\sum_{d'} \exp(\ell_{d'})}.
    \label{eq:loc_loss}
\end{equation}
The total training objective combines the global report--volume contrastive loss
$\mathcal{L}_{\text{global}}$, the prompt loss $\lambda \, \mathcal{L}_{\text{prompt}}$ (Eq.~\ref{eq:promptloss}), and the localization loss $\beta \, \mathcal{L}_{\text{loc}}$.

\begin{table}[ht]
\caption{
Text-to-image retrieval on CT-RATE, Rad-ChestCT (RC) and Merlin validation sets.
Training datasets: A: RefCT (ours, 78k), B: CT-RATE~\cite{ctrate} (47k), C: Merlin~\cite{merlin} (15k), D: INSPECT~\cite{inspect} (19k). $^*$fVLM additionally trains on MedVL-CT69~\cite{fvlm} (272k).
We highlight \bestci{values within the confidence interval of the highest score} separately for the comparison of our main model with baselines (top) and our model with our ablations (bottom).
}
{
\setlength{\tabcolsep}{1pt}
\label{tab:retrieval}
\begin{tabular}{lccccrcrcrcrcrcrc}
\toprule
& \multicolumn{4}{c}{Data} & \multicolumn{6}{c}{CT-RATE} & \multicolumn{2}{c}{RC} & \multicolumn{4}{c}{Merlin R@1} \\
& A & B & C & D & \multicolumn{2}{c}{R@10} & \multicolumn{2}{c}{MAP@5} & \multicolumn{2}{c}{AUC} & \multicolumn{2}{c}{AUC} & \multicolumn{2}{c}{Findings} & \multicolumn{2}{c}{Impr.} \\
\midrule
Random Chance &  &  &  &  & 0.3 &  & 0.1 &  & 50.0 &  & 50.0 &  & 0.8 &  & 0.8 &  \\
\midrule
CT-CLIP~\cite{ctrate} &  & $\checkmark$ &  &  & 5.0 &  & 68.3 &  & 73.1 &  & 62.9 &  & --- &  & --- &  \\
MedVista3D-ViT~\cite{medvista3d} &  & $\checkmark$ &  &  & 10.7 &  & --- &  & 77.8 &  & --- &  & --- &  & --- &  \\
MedVista3D-UniMISS &  & $\checkmark$ &  &  & 8.7 &  & --- &  & 78.2 &  & --- &  & --- &  & --- &  \\
Merlin~\cite{merlin} &  &  & $\checkmark$ &  & 2.7 &  & 62.6 &  & 72.8 &  & 64.4 &  & 59.4 &  & 19.4 &  \\
fVLM$^*$~\cite{fvlm} &  & $\checkmark$ &  &  & --- &  & --- &  & 77.8 &  & 68.0 &  & --- &  & --- &  \\
BrgSA~\cite{brgsa} &  & $\checkmark$ &  &  & 22.2 &  & 70.4 &  & 82.9 &  & 74.2 &  & --- &  & --- &  \\
MPS-CT~\cite{mpsct} &  & $\checkmark$ &  &  & 16.3 &  & 71.3 &  & \best{83.8} &  & \best{77.3} &  & --- &  & --- &  \\
SPECTRE~\cite{spectre} &  & $\checkmark$ & $\checkmark$ & $\checkmark$ & 18.2 & \showstd{1.4} & \bestci{76.8} & \showstd{1.7} & 54.3 & \showstd{1.1} & 55.7 & \showstd{0.6} & 44.5 & \showstd{0.7} & 29.2 & \showstd{0.6} \\
\midrule\midrule
RadFinder (Ours) & $\checkmark$ & $\checkmark$ & $\checkmark$ & $\checkmark$ & \best{31.5} & \showstd{1.6} & \best{78.0} & \showstd{1.7} & \best{83.8} & \showstd{0.7} & \bestci{77.0} & \showstd{0.5} & \best{69.0} & \showstd{0.9} & \best{40.3} & \showstd{0.7} \\
\midrule
No localization loss & $\checkmark$ & $\checkmark$ & $\checkmark$ & $\checkmark$ & \bestci{31.5} & \showstd{1.6} & \best{78.2} & \showstd{1.7} & 83.6 & \showstd{0.7} & 76.9 & \showstd{0.5} & 69.3 & \showstd{0.9} & 40.2 & \showstd{0.6} \\
Global loss only & $\checkmark$ & $\checkmark$ & $\checkmark$ & $\checkmark$ & 29.4 & \showstd{1.6} & \bestci{76.6} & \showstd{1.7} & 56.9 & \showstd{1.1} & 62.8 & \showstd{0.6} & \best{70.7} & \showstd{0.7} & \best{42.2} & \showstd{0.7} \\
Prompt loss only & $\checkmark$ & $\checkmark$ & $\checkmark$ & $\checkmark$ & 5.6 & \showstd{0.8} & \bestci{77.9} & \showstd{1.7} & \best{84.8} & \showstd{0.7} & \best{78.6} & \showstd{0.5} & 13.1 & \showstd{0.6} & 10.9 & \showstd{0.4} \\
\midrule
RefCT dataset & $\checkmark$ &  &  &  & 26.3 & \showstd{1.6} & \bestci{76.7} & \showstd{1.7} & 80.4 & \showstd{0.8} & 77.7 & \showstd{0.6} & 60.5 & \showstd{0.9} & 35.8 & \showstd{0.7} \\
Public datasets, no loc. &  & $\checkmark$ & $\checkmark$ & $\checkmark$ & 26.9 & \showstd{1.6} & \bestci{77.1} & \showstd{1.7} & 83.2 & \showstd{0.7} & 73.6 & \showstd{0.6} & 67.5 & \showstd{0.9} & 40.2 & \showstd{0.8} \\
\bottomrule
\end{tabular}}
\end{table}

\begin{table}[ht]
\centering
\caption{Results for snippet localization on the RefCT test set. 
$^*$SigLIP2 finetuned on RefCT snippet--slice pairs via contrastive loss.
}
\label{tab:localization}
{

\begin{tabular}{lrcrcrcrc}
\toprule
 & \multicolumn{2}{c}{MAE (mm)} & \multicolumn{2}{c}{$<$6mm} & \multicolumn{2}{c}{$<$18mm} & \multicolumn{2}{c}{$<$30mm} \\
\midrule
Random slice & 126.9 & \showstd{5.7} & 4.7 & \showstd{1.1} & 11.8 & \showstd{1.6} & 17.8 & \showstd{2.0} \\
Middle slice & 95.8 & \showstd{3.6} & 4.6 & \showstd{1.2} & 13.0 & \showstd{1.8} & 20.8 & \showstd{2.2} \\
\midrule
CLIP ViT-B/32~\cite{clip} & 124.6 & \showstd{5.5} & 3.2 & \showstd{0.8} & 9.4 & \showstd{1.4} & 14.8 & \showstd{1.7} \\
BiomedCLIP~\cite{biomedclip} & 86.6 & \showstd{5.2} & 8.3 & \showstd{1.3} & 20.2 & \showstd{1.9} & 30.5 & \showstd{2.2} \\
MedSigLIP-448~\cite{medgemma} & 75.6 & \showstd{5.0} & 9.7 & \showstd{1.4} & 26.2 & \showstd{2.1} & 40.1 & \showstd{2.3} \\
SigLIP2~\cite{siglip2} & 82.4 & \showstd{4.7} & 7.8 & \showstd{1.3} & 19.3 & \showstd{1.8} & 31.5 & \showstd{2.3} \\
SigLIP2 finetuned$^*$
& 67.0 & \showstd{4.9} & 17.4 & \showstd{1.8} & 35.8 & \showstd{2.4} & 48.7 & \showstd{2.3} \\
\midrule\midrule
RadFinder (Ours) & \bestci{36.3} & \showstd{2.9} & \bestci{20.3} & \showstd{2.0} & \best{45.3} & \showstd{2.5} & \bestci{61.8} & \showstd{2.4} \\
\midrule
Loc. loss only & \best{36.0} & \showstd{2.9} & \bestci{19.9} & \showstd{2.0} & \bestci{45.2} & \showstd{2.5} & \best{62.7} & \showstd{2.4} \\
\bottomrule
\end{tabular}}
\end{table}

\section{Results}

\myparagraph{Implementation details}
We fine-tune for 10 epochs with AdamW (lr $2{\times}10^{-4}$, 1 epoch linear warmup, cosine decay to $1{\times}10^{-6}$) and an effective batch size of 4096 on a single NVIDIA H100 (96\,GB, 32h). We use loss weights $\lambda=8$ and $\beta=1$.
We report 95\% bootstrap confidence intervals ($B{=}10{,}000$) over test set samples.

\myparagraph{Datasets and protocols}
We evaluate on three external benchmarks.
CT-RATE~\cite{ctrate} provides 1564 reports paired with 3039 volumes (multiple reconstructions per study). We report text-to-image retrieval (R@10) over the full 3039-volume pool, volume-to-volume retrieval (MAP@5) using disease-label IoU as relevance, and disease classification (AUC) over 18 pathology classes.
For the disease class prediction, we use the 7~prompt templates (e.g., ``\{a\} is present.'') proposed by CT-RATE and classify by softmax over cosine similarities between the volume embedding and the averaged positive/negative prompt embeddings.
Rad-ChestCT~\cite{radchestct} provides 3630 chest CT scans with disease labels (no reports); we evaluate AUC with the same prompt protocol.
Merlin~\cite{merlin} provides 5125 report--volume pairs; we report R@1 separately for findings and impressions sections (R@1$_\text{find}$, R@1$_\text{impr}$) with the protocol by the original authors, which averages the metric over 100 trials with pools of 128 randomly sampled volumes.
RefCT (internal) is used to evaluate snippet localization at 12\,mm axial resolution. The test set contains 1,564 volumes with 1 snippet--slice pair each.

\myparagraph{Retrieval (Tab.~\ref{tab:retrieval})}
Training on RefCT alone already surpasses all published baselines on CT-RATE retrieval (R@10~26.3 vs.\ 22.2 for BrgSA)---demonstrating strong cross-dataset transfer.
Adding public datasets further improves retrieval to R@10~31.5.
On the Merlin dataset, our model achieves R@1$_\text{find}$~69.0, surpassing the original Merlin model (59.4).

\myparagraph{Disease classification (Tab.~\ref{tab:retrieval})}
Without prompt-based training, our model achieves only AUC~${\sim}$57 on CT-RATE---comparable to SPECTRE and well below methods that incorporate disease knowledge.
Adding prompts yields AUC~83.8 on CT-RATE and 77.0 on Rad-ChestCT, matching MPS-CT~\cite{mpsct} (83.8 / 77.3) within confidence bounds.
Training on RefCT alone already reaches AUC~80.4 / 77.7, showing that disease knowledge transfers from our internal data.
Conversely, training with prompt loss only yields the highest classification AUC (84.8) but sacrifices retrieval (R@10~5.6), as expected when the model sees only binary disease labels without free-text reports.
The combined model achieves the best of both: strong retrieval and competitive classification without tradeoff.

\myparagraph{Localization (Tab.~\ref{tab:localization})}
Naive baselines (middle slice, random slice) achieve MAE~96 and 127\,mm.
Pretrained 2D VLMs reduce this to 87\,mm (BiomedCLIP~\cite{biomedclip}) and 76\,mm (MedSigLIP~\cite{medgemma}); finetuning SigLIP2~\cite{siglip2} on our snippet--slice pairs reaches 67\,mm.
RadFinder achieves MAE~36.3\,mm (20.3\% within 6\,mm), nearly halving the error of the best baseline.
Training with localization loss alone yields comparable results (MAE~36.0\,mm), confirming that the global objectives do not interfere with local grounding.
Conversely, adding localization to the full model leaves retrieval and classification unchanged within confidence bounds (Tab.~\ref{tab:retrieval}, RadFinder vs.\ No loc.), yielding a single unified model for all three tasks.

\section{Conclusion}

We presented RadFinder, a 3D CT vision--language model that combines contrastive pretraining on reports with prompt-based disease label supervision, achieving state-of-the-art retrieval and competitive disease classification on external benchmarks.
We additionally showed that snippet--slice references mined from radiology reports provide effective local supervision for intra-scan localization, and that this task can be trained jointly with global objectives in a single unified model without degrading retrieval or classification.

\myparagraph{Limitations}
The localization resolution is limited to 12\,mm along the slice axis, which may be insufficient for precisely localizing small findings. Furthermore, our snippet mining relies on explicit slice references in reports, which limits coverage to institutions where such references are part of the reporting practice.

\myparagraph{Future work}
Promising directions include investigating whether local supervision can also improve global representations, collecting data from additional hospitals to study multi-institutional training, evaluating bilingual inference on the original German reports, and improving localization resolution through finer-grained feature maps.
Coupling the pretrained encoder with a language model for grounded report generation is another natural extension.

\begin{credits}
\subsubsection{\ackname}
This research was funded by the 
Deutsche Forschungsgemeinschaft (DFG, German Research Foundation)
417962828,
539134284 -- through EFRE (FEIH\_2698644) and the state of Baden-Württemberg,
and 499552394 -- SFB 1597 -- Small Data.
\begin{center}
\includegraphics[width=0.3\textwidth]{BaWue_Logo_Standard_rgb_pos.png} ~~~ 
\includegraphics[width=0.3\textwidth]{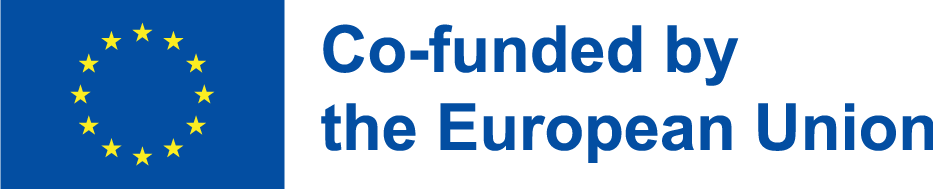}
\end{center} 
\end{credits}

\bibliographystyle{splncs04}
\bibliography{ref}

\end{document}